\documentclass[sigconf, nonacm]{acmart}

\setcopyright{none}

\usepackage{graphicx}
\graphicspath{{./}{./figures/}}

\usepackage{paralist}
\usepackage{multirow}   % enables \multirow
\usepackage{booktabs}   % \toprule \midrulews \bottomrule
\usepackage{amsmath}
\newcommand{\Yes}{\checkmark}     % needs amssymb
\newcommand{\No}{\(\times\)}      % built-in math symbol
\newcommand{\Partial}{\(\triangle\)} % built-in math symbol

\usepackage{hyperref}
\usepackage{color}

\newcommand{\sstitle}[1]{\smallskip\noindent\textbf{#1.\/}}

 \def\Snospace~{\S{}}

\usepackage{xspace}
\newcommand{\toolname}{\texttt{COLA}\xspace}

\begin{document}

\setlength{\belowdisplayskip}{3pt}
\setlength{\belowdisplayshortskip}{3pt}
\setlength{\abovedisplayskip}{3pt}
\setlength{\abovedisplayshortskip}{3pt}

%\title{A Benchmark for Multilingual Text2SQL using Large Language Models}

\title{Multilingual Text-to-SQL: Benchmarking the Limits of Language Models with Collaborative Language Agents}

\author{Khanh Trinh Pham, Thu Huong Nguyen, Jun Jo, Quoc Viet Hung Nguyen, Thanh Tam Nguyen}
\affiliation{
\institution{Griffith University, Australia}
\country{\vspace{.8em}}
}

\begin{abstract}
Text-to-SQL enables natural access to databases, yet most benchmarks are English-only, limiting multilingual progress. We introduce MultiSpider 2.0, extending Spider 2.0 to eight languages (English, German, French, Spanish, Portuguese, Japanese, Chinese, Vietnamese). It preserves Spider 2.0's structural difficulty while adding linguistic and dialectal variability, demanding deeper reasoning for complex SQL. On this benchmark, state-of-the-art LLMs (such as DeepSeek-R1 and OpenAI o1) reach only 4\% execution accuracy when relying on intrinsic reasoning, versus 60\% on MultiSpider 1.0. Therefore, we provide a collaboration-driven language agents baseline that iteratively refines queries, improving accuracy to 15\%. These results reveal a substantial multilingual gap and motivate methods that are robust across languages and ready for real-world enterprise deployment. Our benchmark is available at \url{https://github.com/phkhanhtrinh23/Multilingual_Text_to_SQL}.
\end{abstract}

\maketitle    

\section{Introduction}
Text-to-SQL translates natural language into executable SQL, enabling non-experts to query relational data in QA systems and task-oriented interfaces \cite{li2014constructing}. Despite strong progress with LLMs \cite{chen2021evaluating,austin2021program}, most benchmarks (Spider~1.0 \cite{yu2018spider}, Spider~2.0 \cite{lei2024spider}, BIRD \cite{li2024can}, WikiSQL \cite{zhong2017seq2sql}) remain English-centric, and methods often overfit to monolingual assumptions \cite{wang2023mac,zhang2023act,pourreza2024din,gao2023text,li2023resdsql,dong2023c3,li2024pet,biswal2024text2sql}. MultiSpider~1.0 \cite{dou2023multispider} opened a multilingual path, but it was benchmarked using small language models (LMs) (mBART \cite{liu2020mbart}, XLM-R \cite{conneau-etal-2020-xlm-r}) and just delivers non-industrial schemas.

This paper introduces \emph{MultiSpider~2.0}, extending Spider~2.0's enterprise-grade complexity to eight languages: English (en), German (de), French (fr), Spanish (es), Portuguese (pt), Japanese (ja), Chinese (zh), and Vietnamese (vi). According to \autoref{tab:functional-comparison}, it preserves large, cross-domain schemas and compositional SQL hardness while adding linguistic and dialectal variation, yielding a realistic multilingual setting. On this benchmark, state-of-the-art reasoning-first LLMs attain only 4\% execution accuracy (versus approximately 60\% on MultiSpider~1.0), exposing a pronounced multilingual gap and motivating method development. 

Our contributions are as follows:
\begin{compactitem}
    \item \emph{Extensible benchmark:} We prepare \textit{MultiSpider~2.0}, the first \emph{enterprise-scale, multilingual, dialect-aware} Text-to-SQL benchmark spanning eight languages, with fine-grained slices for schema complexity and compositional reasoning.
	\item \emph{New method:} We propose a \emph{collaborative language agents} baseline that iteratively decomposes, executes, and refines SQL; it improves execution accuracy to 14\% without task-specific finetuning. This baseline is provided to stimulate future method development; its design and results are \emph{reported separately from benchmark contributions}.
    \item \emph{Comprehensive evaluation:} We systematically evaluate strong LLMs on MultiSpider~2.0 and quantify the multilingual gap, highlighting failure cases with schema linking, multi-hop joins, and nested queries.
\end{compactitem}

\begin{table}[t]
\centering
\scriptsize
\setlength{\tabcolsep}{2pt} % Adjusted column separation for the new column
\caption{Compact functional comparison of Text-to-SQL benchmarks. 
\emph{Legend}: \Yes\ = fully supported; \Partial\ = partially supported/limited; \No\ = not supported. 
``Enterprise'' = large, real-world schemas; ``Compositional'' = nested/subqueries and multi-hop joins; 
``Dialect'' = cross-lingual/dialectal variation; ``Diagnostic'' = fine-grained capability slices;
``Multilingual'' = supports non-English languages.
The notation ``(kL)'' after a benchmark name shows the number of languages.}
\label{tab:functional-comparison}
\begin{tabular}{@{}p{0.28\linewidth} c c c c c@{}}
\toprule
\textbf{Benchmark (\#L)} & \textbf{Enterprise} & \textbf{Compositional} & \textbf{Dialect} & \textbf{Diagnostic} & \textbf{Multilingual} \\
\midrule
WikiSQL (1L) \cite{zhong2017seq2sql}             & \No      & \Partial & \No      & \No        & \No \\
Spider~1.0 (1L) \cite{yu2018spider}              & \No      & \Yes     & \No      & \Partial   & \No \\
BIRD (1L) \cite{li2024can}                        & \Partial & \Yes     & \No      & \Partial   & \No \\
Spider~2.0 (1L) \cite{lei2024spider}              & \Yes      & \Yes     & \Yes      & \Yes   & \No \\
MultiSpider~1.0 (7L) \cite{dou2023multispider}   & \No      & \Yes & \Partial & \Yes        & \Yes \\
\textbf{MultiSpider~2.0 (8L)}  & \Yes     & \Yes     & \Yes     & \Yes       & \Yes \\
\bottomrule
\end{tabular}
\end{table}

% \begin{figure*}[h]
%     \centering
%     \includegraphics[width=0.9\linewidth]{multispider.drawio.png}
%     \caption{Overview of the collaborative multi-agent pipeline (CMAS).}
%     \label{fig:cmaspipeline}
% \end{figure*}

\section{MultiSpider 2.0}
\label{sec:multispider2.0}
% Ensure \sstitle exists (no-op if already defined)
\providecommand{\sstitle}[1]{\vspace{0.35em}\noindent\textbf{#1.}\ }

\subsection{Dataset Collection}

MultiSpider~2.0 extends Spider~2.0 to a multilingual, enterprise setting. We curate real databases from cloud marketplaces (e.g., BigQuery public datasets, Snowflake Marketplace) and adopt Spider~2.0 inclusion rules: each database must have $\geq 200$ columns \emph{or} a nested schema \cite{lei2024spider}.  
The release aggregates 5,056 NL--SQL pairs across 200 enterprise databases and covers 8 languages (en, de, fr, es, pt, ja, zh, vi), enabling rapid iteration while preserving task difficulty. All splits ship with gold SQL to support fully local evaluation.

\begin{table*}[!h]
%\vspace{-1em}
  \centering
  \footnotesize
  \caption{Statistics of MultiSpider 2.0 Task Features.}
  \label{tab:spider_stats}
  \vspace{-1em}
  \begin{tabular}{@{}l|l|c@{}}
    \toprule
    \textbf{Task Feature} & \textbf{Sub-feature} & \textbf{Number (\% of Total)} \\
    \midrule
    \multirow{2}{*}{\textbf{Overall Statistics}} & Total Examples & \textbf{5056 (100\%)} \\
    & Total Languages & 8 \\
    \midrule
    \multirow{8}{*}{\textbf{Linguistic Distribution}} & English (en) & 632 (12.5\%) \\
    & German (de) & 632 \\
    & French (fr) & 632 \\
    & Spanish (es) & 632 \\
    & Portuguese (pt) & 632 \\
    & Japanese (ja) & 632 \\
    & Chinese (zh) & 632 \\
    & Vietnamese (vi) & 632 \\
    \midrule
    \multirow{3}{*}{\textbf{Dataset Complexity}} & Easy (< 80 tokens) & 1280 (25.32\%) \\
    & Medium (80-160 tokens) & 2232 (44.15\%) \\
    & Hard (> 160 tokens) & 1544 (30.54\%) \\
    \midrule
    \multirow{4}{*}{\textbf{Query and Schema Complexity}} & With Multiple Schemas & 1120 (22.15\%) \\
    & With Nested Schemas & 936 (18.51\%) \\
    & With Partition Tables & 432 (8.54\%) \\
    & With Functions & 3792 (75.00\%) \\
    \midrule
    \multirow{3}{*}{\textbf{Database Dialects}} 
    & With Bigquery  & 1712 (33.86\%) \\
    & With Snowflake & 1584 (31.33\%) \\
    & With SQLite    & 1760 (34.81\%) \\
    \bottomrule
  \end{tabular}
\end{table*}

Following Lei et al. \cite{lei2024spider}, \autoref{tab:spider_stats} details the compositional statistics of the MultiSpider 2.0 benchmark, revealing the foundations of its increased difficulty. 
Compared to MultiSpider \cite{dou2023multispider}, this benchmark maintains a balanced distribution of complexity based on query length, with a significant portion of tasks classified as Medium (44.15\%) and Hard (30.54\%). 
This ensures that models are evaluated on non-trivial problem-solving capabilities. MultiSpider 2.0 mirrors contemporary enterprise data environments by moving beyond a single SQL dialect. 
It incorporates a wide array of modern data warehouse syntaxes, with strong representation from BigQuery (33.86\%), Snowflake (31.33\%), and SQLite (34.81\%). 
This diversity tests a model's robustness to dialectal variations.

\subsection{Translation Procedure and Challenges}
\label{subsec:translation_procedure}

\begin{figure}[!h]
  \centering
  \includegraphics[width=1\linewidth]{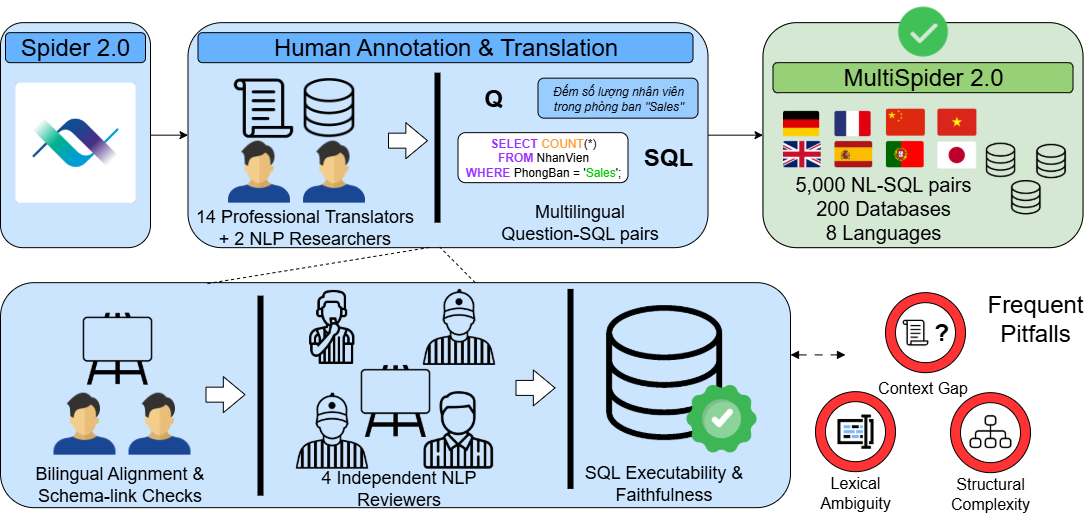}
  \caption{Dataset Construction Pipeline.}
  \label{fig:dataset_construction}
\end{figure}

\sstitle{Pipeline}
As in \autoref{fig:dataset_construction}, translations are produced by a team of 14 professional translators (two per non-English language), plus two NLP researchers overseeing English. We localize not only the natural-language questions but also the database for eight target languages -- English, German, French, Spanish, Portuguese, Japanese, Chinese, and Vietnamese -- so that each language has a self-contained schema and value inventory aligned to the source.

A multi-pass process follows: (i) bilingual alignment and schema-link checks for both questions and schema/value dictionaries; (ii) construction of per-language, localized database snapshots in which table/column names, enumerations, and reference strings are translated while canonical IDs are preserved; (iii) four independent NLP reviewers verify SQL executability and faithfulness against the localized schemas; and (iv) cross-lingual equivalence checks to ensure each localized instance is semantically identical to the original. This process yields consistent schema grounding across languages.

\sstitle{Challenges}
According to \autoref{tab:example_explanation_table}, we observe three recurrent error types that informed our QA rubrics and diagnostic tags:
\begin{compactitem}
    \item \emph{Context gaps} (domain terms or schema elements under-specified), leading to missing column constraints.
    \item \emph{Lexical ambiguity} (polysemy, named products/entities), harming string matching and join disambiguation.
    \item \emph{Structural complexity} (nested logic, multi-hop joins), causing operator/column misalignment and ranking errors.
\end{compactitem}

\sstitle{Our approach}
To mitigate these, we provide a bilingual schema dictionary with round-trip checks and execute smoke tests on each localized snapshot. \autoref{tab:example_explanation_table} shows how targeted fixes (explicit device categories; retaining canonical product names; precise type filters) preserve the intended SQL semantics.

\begin{table*}[!h]
    \centering
    \footnotesize
    \caption{Examples of common annotation challenges and resolutions for translating English text-to-SQL prompts and database fields to the eight target languages (illustrated here with English $\rightarrow$ Vietnamese).}
    \label{tab:example_explanation_table}
    \begin{tabular}{p{3cm}|p{9.5cm}|p{4.5cm}}
    \hline
    \multicolumn{1}{c|}{\textbf{Challenge}} & \multicolumn{1}{c|}{\textbf{Example (en $\rightarrow$ vi)}} & \multicolumn{1}{c}{\textbf{Fix / Principle}} \\
    \hline
    \textbf{Contextual knowledge gap} & \textbf{Q\_en:} For each visitor, days between the first transaction and the first visit in Feb 2017; also report the device category. \newline
    \textbf{Q\_vi (bad):} So ngay giua giao dich dau tien va lan truy cap dau tien trong thang 2/2017 cho moi khach. \newline
    \textbf{Q\_vi (fix):} Cho moi khach truy cap, tinh so ngay giua giao dich dau tien va lan truy cap dau tien trong thang 2/2017, \emph{theo phan loai thiet bi (may tinh, di dong)}.
    & Make implicit fields explicit; map to schema columns (e.g., \texttt{device.deviceCategory}). Avoid vague additions not grounded in schema. \\
    \hline
    \textbf{Lexical ambiguity} & \textbf{Q\_en:} Top-selling product among customers who bought ``YouTube Men's Vintage Henley'' in July 2017 (exclude that product). \newline
    \textbf{Q\_vi (bad):} Tim san pham ban chay cho khach mua ``Ao Henley Vintage nam Youtube''. \newline
    \textbf{Q\_vi (fix):} Tim san pham ban chay nhat cho khach da mua \emph{``YouTube Men's Vintage Henley''} trong 07/2017, \emph{va loai tru chinh san pham nay khoi ket qua}.
    & Keep canonical named entities in quotes; do not translate brand/model strings. State exclusion explicitly to match filter logic. \\
    \hline
    \textbf{Structural complexity} & \textbf{Q\_en:} Which month has the greatest number of \emph{motor vehicle thefts} in 2016? \newline
    \textbf{Q\_vi (bad):} Thang nao co so vu \emph{trom} nhieu nhat nam 2016? \newline
    \textbf{Q\_vi (fix):} Thang nao co so vu \emph{trom xe co gioi} nhieu nhat nam 2016?
    & Use the correct subtype/category in filters (e.g., \texttt{'MOTOR VEHICLE THEFT'} not \texttt{'THEFT'}); verify label taxonomy before counting. \\
    \hline
    \end{tabular}
\end{table*}

\section{Benchmark Settings}
\label{sec:benchmark_settings}

\begin{figure}[!htbp]
  \centering
  \includegraphics[width=1.0\linewidth]{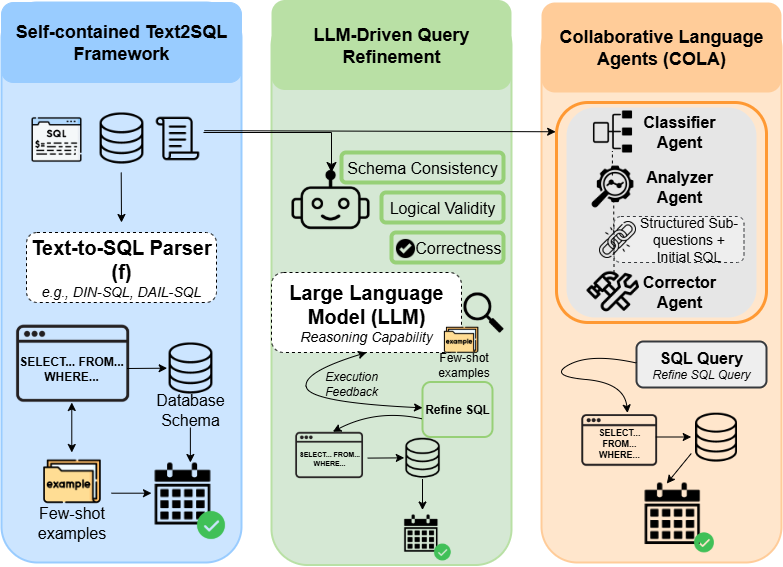}
  \caption{Benchmark Settings for MultiSpider 2.0.}
  \label{fig:benchmark_settings}
\end{figure}

As in \autoref{fig:benchmark_settings}, our experiments are conducted in three distinct settings: (i) traditional Text-to-SQL generation, (ii) a paradigm utilizing the built-in reasoning capabilities of LLMs, and (iii) our proposed method, called \textit{Collaborative Language Agents (\toolname)}. These setups allow us to evaluate performance under both direct query generation and LLM-assisted refinement.

\subsection{Self-Contained Text-to-SQL Framework}
\label{subsec:self_contained_text2sql_task}
Given a database schema \( D \), a natural language question \( Q \), and auxiliary documentation \( E \), a parser \( f(\cdot) \) generates the SQL query \( S = f(Q, D, E \mid \theta) \), where \( \theta \) represents the model's parameters. In the self-contained setting, a database schema, a natural language question, and auxiliary documentation are provided as inputs. A pre-trained Text-to-SQL parser generates the corresponding SQL query, which is then executed against the database. Performance is evaluated based on the accuracy of the generated SQL query compared to the ground truth. We evaluate several state-of-the-art Text-to-SQL methods on MultiSpider 2.0 and MultiSpider 1.0, including prompting-based approaches like DIN-SQL \cite{pourreza2024din}, DAIL-SQL \cite{gao2023text}, TAG-Bench \cite{biswal2024text2sql}, RESDSQL \cite{li2023resdsql}, C3SQL \cite{dong2023c3}, PETSQL \cite{li2024pet}, and CHESS \cite{talaei2024chess}. Among them, DAIL-SQL \cite{gao2023text} achieves the highest performance on MultiSpider 2.0, excelling in context-aware prompting and SQL dialect adaptation. To ensure fair comparison, we optimize prompt structures across all methods by incorporating sampled cell values, external knowledge from the additional documentation, and database-specific SQL constraints. These upgrades enhance model adaptability to multilingual and complex query generation. 

\subsection{LLM-Driven Query Refinement}
\label{subsec:llm_refinement}
Instead of relying on prompting techniques and auxiliary documentation as in Section \ref{subsec:self_contained_text2sql_task}, this setting leverages the inherent reasoning capabilities of LLMs to refine SQL queries iteratively on the-fly, based on process-supervised manner \cite{lightman2023let}. Given an initial query generated from the self-contained setting, an LLM assesses its correctness based on schema consistency, logical validity, and execution feedback, as in famous works of OpenAI-o1 \cite{lightman2023let} and DeepSeek-R1 \cite{guo2025deepseek}, until the final optimized query is obtained. This method effectively utilizes the chain-of-thought and self-correction abilities of modern LLMs to enhance query generation accuracy in multilingual Text-to-SQL tasks.

\subsection{Collaborative Language Agents (COLA)}
\label{subsec:proposal}

The LLM-Driven method outlined in Section \ref{subsec:llm_refinement} facilitates iterative query refinement by utilizing the inherent reasoning abilities of LLMs. However, depending on model-generated revisions and process supervision can result in \textit{extended query generation times} and \textit{higher computational costs}, due to the "extended test-time computation" \cite{guo2025deepseek}. This issue becomes more pronounced when dealing with complex database structures or ambiguous natural language queries. Additionally, relying on a single model can lead to suboptimal results due to a lack of revisions and feedback. To mitigate these limitations, we introduce \textit{Collaborative Language Agents (\toolname)}, as depicted in~\autoref{fig:benchmark_settings}. COLA leverages LLM-based intelligent agents, each tailored to specific tasks, to improve the accuracy of Text-to-SQL parsing. The framework includes three main agents:

\begin{itemize}
    \item \textit{Classifier:} Identifies and partitions large databases into smaller, more relevant sub-databases. This reduces noise from irrelevant information, improving query precision.  
    \item \textit{Analyzer:} Decomposes complex user queries into structured sub-questions and generates SQL queries through chain-of-thought reasoning.  
    \item \textit{Corrector:} Executes SQL queries, analyzes feedback, and refines queries iteratively to correct syntax and logic errors.  
\end{itemize}

\section{Experiments}
\label{sec:experiments}

\subsection{Experimental Setup}

\sstitle{Evaluation Protocol}
We evaluate models in three distinct settings to comprehensively assess their capabilities. 
(i)~\emph{Self-contained Parsers}: Specialized text-to-SQL models that rely on sophisticated prompting strategies, such as DIN-SQL, DAIL-SQL, and RESDSQL, all paired with a GPT-4o backbone.
(ii)~\emph{Reasoning-only LLMs}: State-of-the-art LLMs (Gemini~1.5~Pro, OpenAI-o1-1217, DeepSeek-R1 variants) are prompted to directly generate the SQL query from the natural language question and database schema. This tests their intrinsic text-to-SQL capabilities.
(iii)~\emph{Collaborative Language Agents (COLA)}: Our proposed method, which employs specialized agents for query analysis, decomposition, execution, and correction. We test COLA with each of the reasoning-only LLMs as a backbone to demonstrate its modularity and effectiveness.
All experiments are conducted on both MultiSpider~1.0 and MultiSpider~2.0, using identical schema serialization, and few-shot examples to ensure a fair comparison.

\sstitle{Metrics}
We evaluate performance using two standard metrics: Exact Matching (EM) and Execution Accuracy (EX). Given a predicted query $\hat{S}_j$ and a gold query $S^*_j$ for an example $j$, the metrics are defined as:
\begin{align}
\mathrm{EM}=\tfrac{1}{m}\sum_{j=1}^m \mathbf{1}[\mathrm{norm}(\hat{S}_j)=\mathrm{norm}(S^*_j)] \\
\mathrm{EX}=\tfrac{1}{m}\sum_{j=1}^m \mathbf{1}[\mathrm{Exec}(\hat{S}_j)=\mathrm{Exec}(S^*_j)]
\end{align}
where $\mathrm{norm}(\cdot)$ standardizes the SQL query text and $\mathrm{Exec}(\cdot)$ returns the result set from executing the query on the database. We also report Pass@$N$ for $N \in \{5, 10, 20\}$, where at least one of the top-$N$ generated queries is correct.

\subsection{Main Results}

\begin{figure}[!h]
  \centering
  \includegraphics[width=0.7\linewidth]{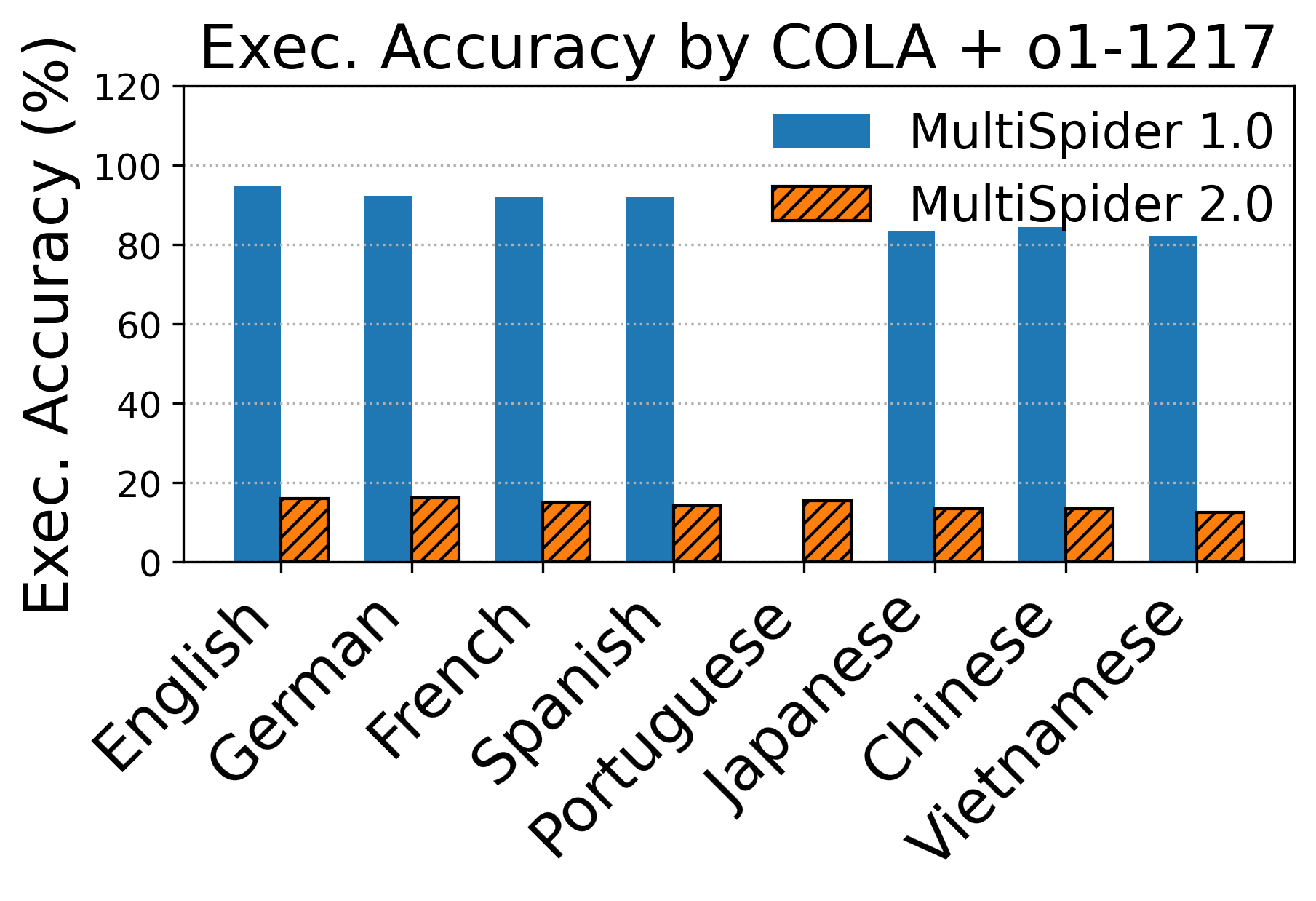}
  \caption{Execution Accuracy (EX) comparison between MultiSpider 1.0 and MultiSpider 2.0 using the COLA + OpenAI-o1-1217 model. The drastic drop in performance highlights the increased difficulty of MultiSpider 2.0 across all languages.}
  \label{fig:ex_results}
\end{figure}

\begin{figure}[!h]
  \centering
  \includegraphics[width=0.7\linewidth]{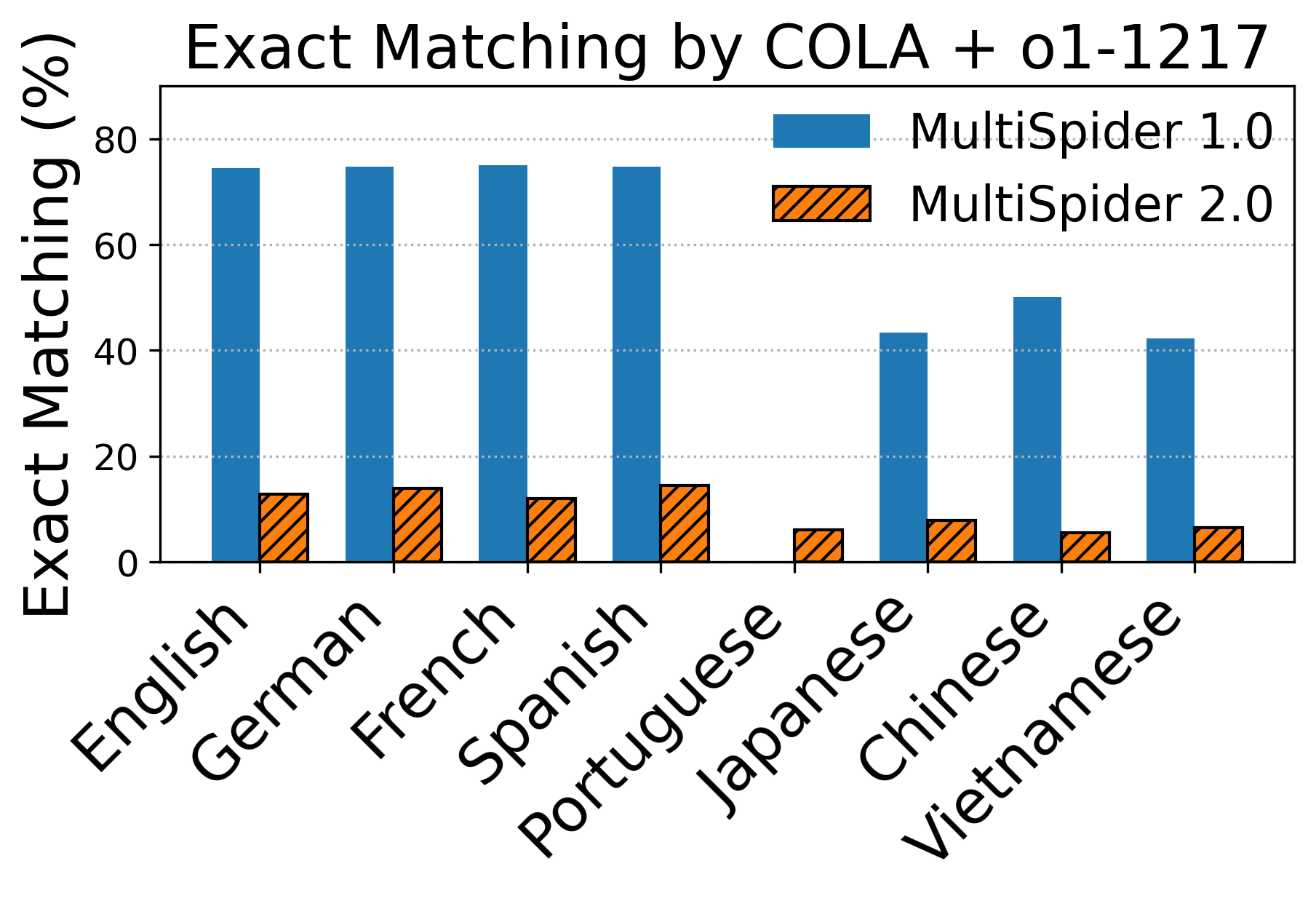}
  \caption{Exact Matching (EM) comparison between MultiSpider 1.0 and MultiSpider 2.0 for the COLA + OpenAI-o1-1217 model. Similar to EX, EM scores fall sharply on the new benchmark, indicating syntactic and structural complexity.}
  \label{fig:em_results}
\end{figure}

\sstitle{MultiSpider 2.0 Presents a Formidable Challenge}
Our primary finding is that MultiSpider 2.0 is substantially more difficult than its predecessor. As illustrated in \autoref{fig:ex_results} and \autoref{fig:em_results}, even our strongest model, COLA with an OpenAI-o1-1217 backbone, experiences a dramatic performance degradation. On MultiSpider 1.0, this model achieves Execution Accuracy (EX) scores above 90\% for several European languages. However, on MultiSpider 2.0, its accuracy plummets to a range of 12--16\%. This performance drop of over 75 percentage points underscores the new challenges introduced by complex enterprise schemas, linguistic diversity, and difficult query compositions in MultiSpider 2.0.

% Execution Accuracy table

% (use your existing Table \ref{tab:execution_accuracy} here)

\begin{table*}[!h]
\centering
\footnotesize
\caption{Execution accuracy (\%) across languages. The top row for each method corresponds to MultiSpider 1.0, while the bottom row corresponds to MultiSpider 2.0.}
\label{tab:execution_accuracy}
\begin{tabular}{@{}l|c|c|c|c|c|c|c|c@{}}

\toprule

\multicolumn{1}{c|}{\textbf{Methods}} & \textbf{en} & \textbf{de} & \textbf{fr} & \textbf{es} & \textbf{pt} & \textbf{ja} & \textbf{zh} & \textbf{vi} \\

\midrule

\multirow{2}{*}{DIN-SQL + GPT-4o} & 80.55 & 80.11 & 80.06 & 80.48 & -- & 78.27 & 73.14 & 70.51 \\

& 12.98 & 12.49 & 13.21 & 12.87 & 9.1 & 8.99 & 8.01 & 9.13 \\

\hline

\multirow{2}{*}{DAIL-SQL + GPT-4o} & 84.25 & 83.32 & 83.76 & 83.45 & -- & 72.18 & 78.68 & \textbf{82.37} \\

& 12.3 & 12.13 & 13.05 & 14.76 & 10.94 & 8.43 & 9.04 & 8.36 \\

\hline

\multirow{2}{*}{TAG-Bench + GPT-4o} & 81.53 & 81.4 & 81.04 & 81.34 & -- & 70.25 & 76.83 & 73.89 \\

& 12.56 & 13.3 & 14.01 & 13.78 & 9.24 & 11.86 & 8.31 & 10.26 \\

\hline

\multirow{2}{*}{RESDSQL + GPT-4o} & 81.55 & 80.85 & 80.47 & 80.09 & -- & 75.46 & 78.27 & 73.23 \\

& 12.69 & 14.17 & 14.71 & 12.59 & 8.38 & 8.88 & 9.05 & 11.6 \\

\hline

\multirow{2}{*}{C3SQL + GPT-4o} & 81.53 & 81.15 & 81.28 & 81.59 & -- & 79.2 & 78.92 & 77.49 \\

& 13.32 & 13.02 & 14.0 & 14.73 & 8.98 & 8.71 & 9.83 & 11.18 \\

\hline

\multirow{2}{*}{PETSQL + GPT-4o} & 81.29 & 80.69 & 80.33 & 80.96 & -- & 70.68 & 74.54 & 78.73 \\

& 13.22 & 12.02 & 13.05 & 13.54 & 10.38 & 9.79 & 10.79 & 8.22 \\

\hline

\multirow{2}{*}{CHESS + GPT-4o} & 82.18 & 81.95 & 81.7 & 80.61 & -- & 76.37 & 78.7 & 71.65 \\

& 12.09 & 13.38 & 14.15 & 12.14 & 10.92 & 9.32 & 9.55 & 11.94 \\

\hline

\multirow{2}{*}{Gemini 1.5 Pro} & 79.69 & 78.95 & 79.05 & 78.91 & -- & 78.96 & 76.16 & 77.21 \\

& 4.87 & 4.23 & 3.04 & 4.05 & 4.58 & 3.91 & 4.5 & 5.61 \\

\hline

\multirow{2}{*}{OpenAI-o1-1217} & 79.67 & 78.71 & 78.21 & 78.53 & -- & 79.88 & 78.95 & 78.03 \\

& 4.37 & 5.42 & 5.8 & 4.41 & 5.55 & 4.58 & 5.41 & 5.2 \\

\hline

\multirow{2}{*}{DeepSeek-R1-Distill-Qwen-32B} & 79.54 & 79.27 & 79.78 & 76.62 & -- & 76.15 & 77.25 & 76.69 \\

& 5.32 & 5.21 & 5.24 & 5.22 & 4.73 & 4.23 & 5.52 & 4.14 \\

\hline

\multirow{2}{*}{DeepSeek-R1-Distill-Qwen-70B} & 80.01 & 79.15 & 79.37 & 79.68 & -- & 78.4 & 79.65 & 77.1 \\

& 5.83 & 5.61 & 5.46 & 5.47 & 5.33 & 5.64 & 5.89 & 5.22 \\

\hline

\multirow{2}{*}{COLA + Gemini 1.5 Pro} & 89.23 & 86.34 & 87.65 & 86.38 & -- & 81.95 & 82.02 & 80.94 \\

& 15.68 & 15.26 & 14.85 & 14.55 & 13.97 & 12.11 & 13.66 & 12.44 \\

\hline

\multirow{2}{*}{COLA + OpenAI-o1-1217} & \textbf{94.95} & 92.23 & \textbf{91.97} & \textbf{91.95} & -- & 83.45 & 84.46 & 82.15 \\

& 15.92 & \textbf{16.2} & 15.14 & 14.23 & \textbf{15.53} & \textbf{13.43} & 13.43 & 12.49\\

\hline

\multirow{2}{*}{COLA + DeepSeek-R1-Distill-Qwen-32B} & 90.95 & 91.23 & 90.49 & 90.54 & -- & 81.94 & 80.17 & 80.26 \\

& 15.43 & 14.02 & 15.58 & 14.56 & 13.12 & 14.45 & \textbf{15.74} & 13.83 \\

\hline

\multirow{2}{*}{COLA + DeepSeek-R1-Distill-Llama-70B} & 92.24 & \textbf{93.37} & 91.92 & 91.01 & -- & \textbf{86.39} & \textbf{89.21} & 82.04 \\

& \textbf{15.94} & 15.23 & \textbf{15.94} & \textbf{14.88} & 14.83 & 13.37 & 14.78 & \textbf{13.59} \\

\hline

\end{tabular}
\end{table*}

\begin{figure}[!htbp]
  \centering
  \includegraphics[width=0.7\linewidth]{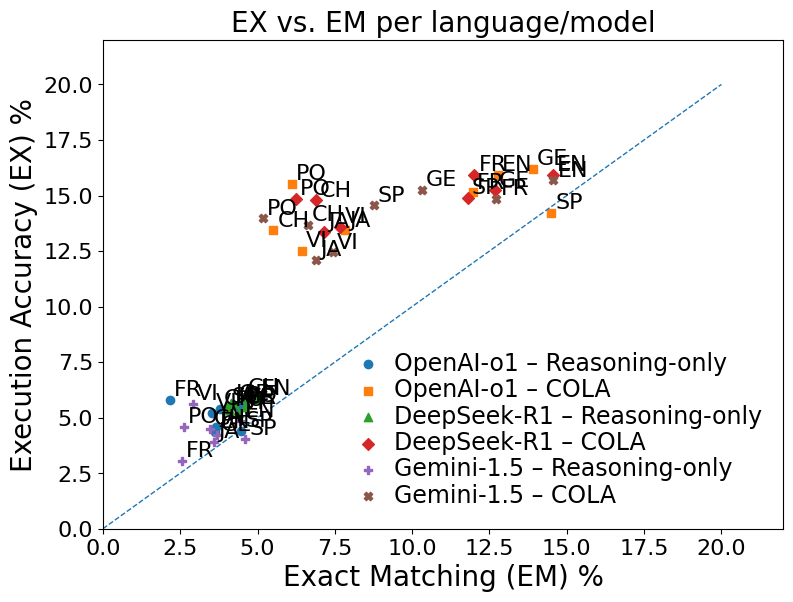}
  \caption{Execution Accuracy (EX) vs. Exact Matching (EM) per language for various models. Most points lie above the $y=x$ line, indicating that models often produce functionally correct queries that do not exactly match the gold standard. COLA-based methods (squares, diamonds, crosses) consistently outperform reasoning-only baselines.}
  \label{fig:ex_vs_em}
\end{figure}

\begin{figure}[!htbp]
  \centering
  \includegraphics[width=0.7\linewidth]{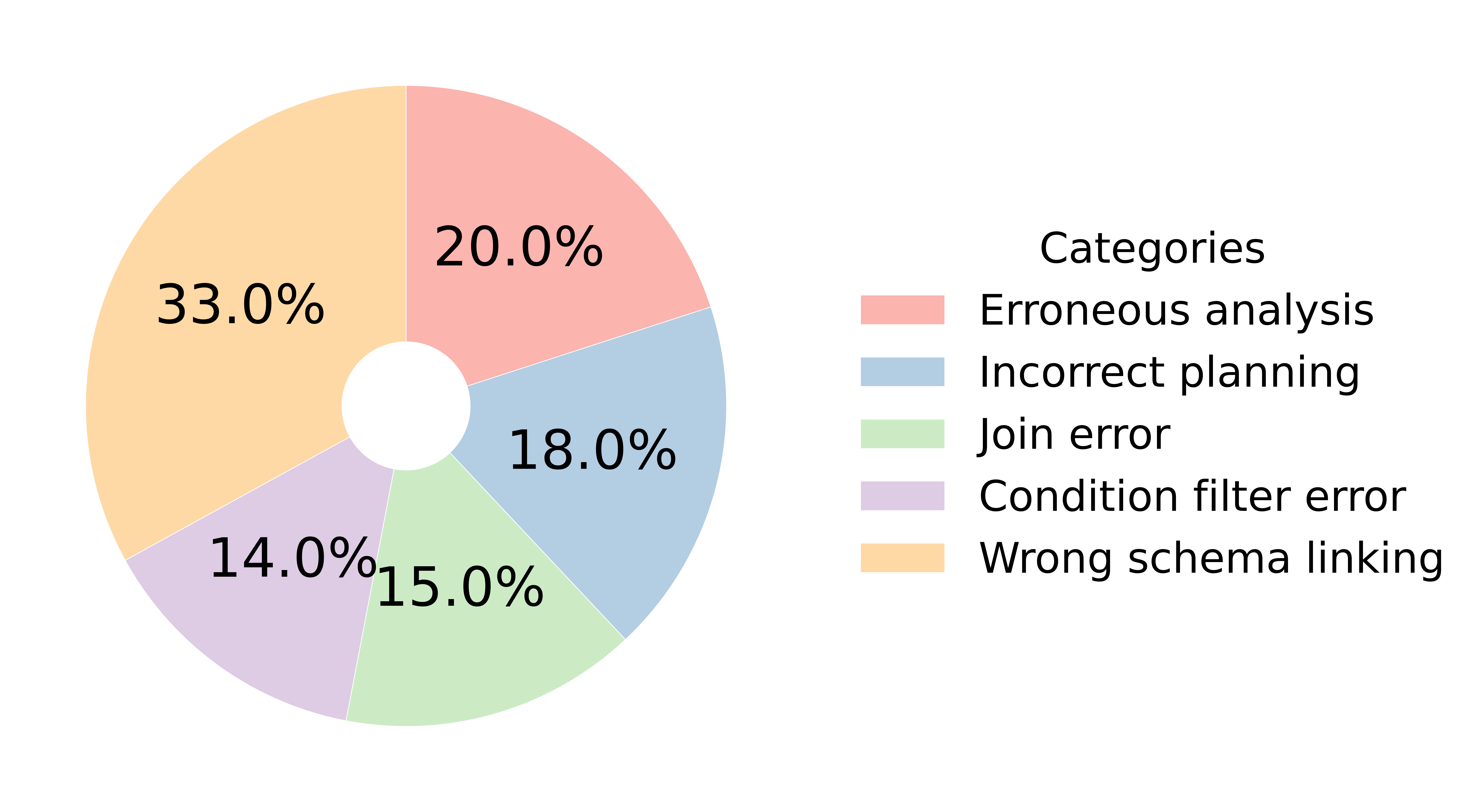}
  \caption{Distribution of error categories on the MultiSpider 2.0 development set. The majority of errors stem from semantic misunderstandings, such as erroneous analysis and wrong schema linking, rather than syntactic issues.}
  \vspace{-1em}
  \label{fig:error_analysis}
\end{figure}

\sstitle{COLA Boosts Performance Across Models and Languages}
Despite the overall difficulty, our proposed COLA framework consistently improves performance over reasoning-only LLM baselines. As detailed in \autoref{tab:execution_accuracy}, standalone models like OpenAI-o1-1217 and Gemini 1.5 Pro achieve only 4--5\% EX on MultiSpider 2.0. By integrating these same models as backbones within COLA, we observe a significant performance lift, with EX scores rising to 13--16\%. This demonstrates that the structured, iterative refinement process of COLA is more effective than the monolithic reasoning of single LLMs for complex, multilingual text-to-SQL tasks. This improvement is also visible in \autoref{fig:ex_vs_em}, where COLA-driven models (e.g., red diamonds, orange squares) consistently plot higher on the EX axis than their reasoning-only counterparts (e.g., blue circles).

\subsection{Analysis and Discussion}

\sstitle{Execution Correctness vs. Syntactic Exactness}
A consistent trend across all models and languages is that Execution Accuracy significantly surpasses Exact Matching (EX $\gg$ EM). This gap is visualized in the scatter plot in \autoref{fig:ex_vs_em}, where nearly all data points lie above the $y=x$ identity line. For instance, on MultiSpider 1.0, COLA + OpenAI-o1-1217 achieves 95.0\% EX in English but only 74.5\% EM (\autoref{tab:exact_matching} and \autoref{tab:execution_accuracy}). This indicates that models frequently generate SQL queries that are semantically and functionally correct but differ syntactically from the gold standard, highlighting the importance of execution-based metrics for practical evaluation. As shown in Tables \ref{tab:exact_matching} and \ref{tab:execution_accuracy}, a significant performance disparity exists across languages.
Models consistently achieve lower scores on non-English languages, particularly for Japanese, Chinese, and Vietnamese.
For example, the performance of COLA + OpenAI-o1-1217 on MultiSpider 1.0 shows an Exact Matching score of 74.46\% in English, which drops to 43.4\% for Japanese and 50.18\% for Chinese. A similar trend is observed in Execution Accuracy.

% --- Tables (unchanged content; references now use \autoref) ---

% Exact Matching table

% (use your existing Table \ref{tab:exact_matching} here)

\begin{table*}[!h]
\vspace{0em}
\centering
\footnotesize
\caption{Exact matching accuracy (\%) across languages. The top row for each method corresponds to MultiSpider 1.0, while the bottom row corresponds to MultiSpider 2.0.}
\label{tab:exact_matching}
\begin{tabular}{@{}l|c|c|c|c|c|c|c|c@{}}
\toprule

\multicolumn{1}{c|}{\textbf{Methods}} & \textbf{en} & \textbf{de} & \textbf{fr} & \textbf{es} & \textbf{pt} & \textbf{ja} & \textbf{zh} & \textbf{vi} \\

\midrule

\multirow{2}{*}{DIN-SQL + GPT-4o} & 72.47 & 68.96 & 65.7 & 65.7 & -- & 54.67 & 58.46 & 58.01\\

& 11.37 & 8.57 & 13.07 & 8.73 & 5.05 & 6.19 & 6.42 & 7.72 \\

\hline

\multirow{2}{*}{DAIL-SQL + GPT-4o} & 74.03 & 72.37 & 71.52 & 67.25 & -- & 44.71 & 45.2 & \textbf{58.54}\\

& 13.81 & 8.92 & 9.49 & 9.89 & \textbf{8.0} & 5.31 & \textbf{7.97} & \textbf{8.99} \\

\hline

\multirow{2}{*}{TAG-Bench + GPT-4o} & 71.91 & 66.67 & 72.22 & 72.5 & -- & 53.73 & 43.75 & 54.55\\

& 8.31 & 13.14 & 12.73 & 9.02 & 5.65 & 6.21 & 5.19 & 7.39\\

\hline

\multirow{2}{*}{RESDSQL + GPT-4o} & 70.78 & 72.01 & 67.53 & 66.3 & -- & 41.83 & 55.81 & 56.03\\

& 11.44 & 9.23 & 12.85 & 12.74 & 7.64 & 6.19 & 5.57 & 7.69\\

\hline

\multirow{2}{*}{C3SQL + GPT-4o} & 67.85 & 69.77 & 68.77 & 68.7 & -- & 51.39 & 52.15 & 53.3\\

& 12.09 & 11.97 & 14.42 & 8.61 & 5.58 & 7.7 & 6.52 & 5.75\\

\hline

\multirow{2}{*}{PETSQL + GPT-4o} & 71.08 & 69.28 & 65.02 & 73.77 & -- & 50.42 & 48.83 & 53.52\\

& 11.86 & 8.11 & 14.33 & 10.08 & 7.46 & 7.65 & 6.74 & 6.0\\

\hline

\multirow{2}{*}{CHESS + GPT-4o} & 68.65 & 65.04 & 71.94 & 66.96 & -- & 50.8 & 52.07 & 56.03\\

& 10.17 & 11.96 & 12.39 & 11.01 & 5.69 & 7.75 & 5.23 & 7.68\\

\hline

\multirow{2}{*}{Gemini 1.5 Pro} & 65.28 & 73.87 & 72.86 & 73.13 & -- & 46.72 & 58.82 & 52.12\\

& 4.44 & 3.69 & 2.55 & 4.59 & 2.62 & 3.58 & 3.46 & 2.9\\

\hline

\multirow{2}{*}{OpenAI-o1-1217} & 67.72 & 68.5 & 68.81 & 66.79 & -- & 48.55 & 56.18 & 49.16\\

& 3.64 & 4.43 & 2.16 & 4.47 & 4.51 & 3.67 & 3.78 & 3.53\\

\hline

\multirow{2}{*}{DeepSeek-R1-Distill-Qwen-32B} & 69.54 & 69.27 & 69.78 & 70.62 & -- & 47.15 & 41.25 & 41.69\\

& 4.45 & 3.14 & 3.94 & 4.52 & 2.64 & 2.79 & 4.02 & 2.43\\

\hline

\multirow{2}{*}{DeepSeek-R1-Distill-Qwen-70B} & 67.0 & 68.15 & 71.37 & 70.68 & -- & 47.4 & 58.65 & 47.1\\

& 4.96 & 4.25 & 4.6 & 4.01 & 4.12 & 4.13 & 4.56 & 4.39\\

\hline

\multirow{2}{*}{COLA + Gemini 1.5 Pro} & 69.7 & 65.59 & 73.81 & 73.58 & -- & 51.17 & 57.22 & 53.79\\

& 14.57 & 10.32 & 12.72 & 8.76 & 5.17 & 6.9 & 6.62 & 7.44\\

\hline

\multirow{2}{*}{COLA + OpenAI-o1-1217} & \textbf{74.46} & \textbf{74.68} & \textbf{74.98} & 74.68 & -- & 43.4 & 50.18 & 42.21\\

& 12.77 & \textbf{13.9} & 11.97 & \textbf{14.5} & 6.11 & \textbf{7.82} & 5.51 & 6.44\\

\hline

\multirow{2}{*}{COLA + DeepSeek-R1-Distill-Qwen-32B} & 70.52 & 71.28 & 69.67 & 70.12 & -- & 48.33 & 49.91 & 50.45\\

& 10.99 & 12.13 & \textbf{14.85} & 10.88 & 5.92 & 6.71 & 6.45 & 7.22\\

\hline

\multirow{2}{*}{COLA + DeepSeek-R1-Distill-Llama-70B} & 72.89 & 73.12 & 71.45 & \textbf{74.78} & -- & \textbf{59.14} & \textbf{59.33} & 52.98\\

& \textbf{14.55} & 12.67 & 12.01 & 11.79 & 6.24 & 7.15 & 6.89 & 7.68\\

\hline

\end{tabular}
\end{table*}

This performance drop might be attributed to several factors:
\begin{compactitem}
    \item \textit{Data Scarcity:} Large language models are pre-trained on English-heavy datasets, leading to a bias towards English syntax and grammar.
    \item \textit{Linguistic Differences:} Non-English languages, such as Japanese and Chinese, have different linguistic structures (e.g., lack of word spacing), which complicates parsing and understanding for models.
    \item \textit{Code-Switching:} The mix of English technical terms (e.g., database schemas) within non-English queries can be challenging for models.
\end{compactitem}
These factors contribute to an average absolute drop of about 6.1\% in accuracy for non-English languages, highlighting the need for multilingual-aware evaluation and model training.

\sstitle{Error Analysis}
To understand the root causes of failure on MultiSpider 2.0, we manually categorized errors on the development set, with the distribution shown in \autoref{fig:error_analysis}. The analysis reveals that failures stem from deep semantic issues rather than simple syntactic mistakes. The two most common error types are \textit{Wrong schema linking} (33.0\%) and \textit{Erroneous analysis} (20.0\%). The former involves incorrectly mapping natural language phrases to the corresponding database columns and tables, while the latter entails misunderstanding complex query constraints or aggregations. Together with \textit{Incorrect planning} (18.0\%), these top categories highlight that models primarily struggle with reasoning about the query's intent relative to the database structure.

\sstitle{Task Difficulty and Sample Efficiency}
The profound difficulty of MultiSpider 2.0 is further confirmed by Pass@$N$ scores (\autoref{tab:pass_n}). Even when given 20 attempts (Pass@20), the top-performing system, COLA + OpenAI-o1-1217, solves only 14.88\% of the problems. This low ceiling suggests that simply generating more answers is insufficient and that more fundamental improvements in reasoning are required. Nonetheless, COLA consistently provides a better Pass@$N$ curve than reasoning-only models, indicating its structured approach is more sample-efficient.

\begin{table}[!h]
\centering
\footnotesize
\caption{Pass@N Scores for MultiSpider 2.0 Datasets (N = 5, 10, 20).}
\label{tab:pass_n}
\begin{tabular}{l|c|c|c}

\hline

\multicolumn{1}{c|}{\textbf{Methods}} & \textbf{Pass@5} & \textbf{Pass@10} & \textbf{Pass@20} \\

\hline

Gemini 1.5 Pro & 6.38\% & 8.22\% & 9.55\% \\

OpenAI-o1-1217 & 6.57\% & 8.16\% & 14.06\% \\

DeepSeek-R1-Distill-Qwen-32B & 6.54\% & 8.22\% & 13.73\% \\

DeepSeek-R1-Distill-Qwen-70B & 6.04\% & 7.71\% & 14.5\% \\

COLA + Gemini 1.5 Pro & 6.3\% & 8.18\% & 12.49\% \\

COLA + OpenAI-o1-1217 & \textbf{7.38\%} & 8.48\% & \textbf{14.88\%} \\

COLA + DeepSeek-R1-Distill-Qwen-32B & 6.24\% & 7.88\% & 12.69\% \\

COLA + DeepSeek-R1-Distill-Qwen-70B & 6.85\% & \textbf{8.66\%} & 14.59\% \\

\hline

\end{tabular}
\end{table}

\begin{figure}[!htbp]
  \centering
  \includegraphics[width=1\linewidth]{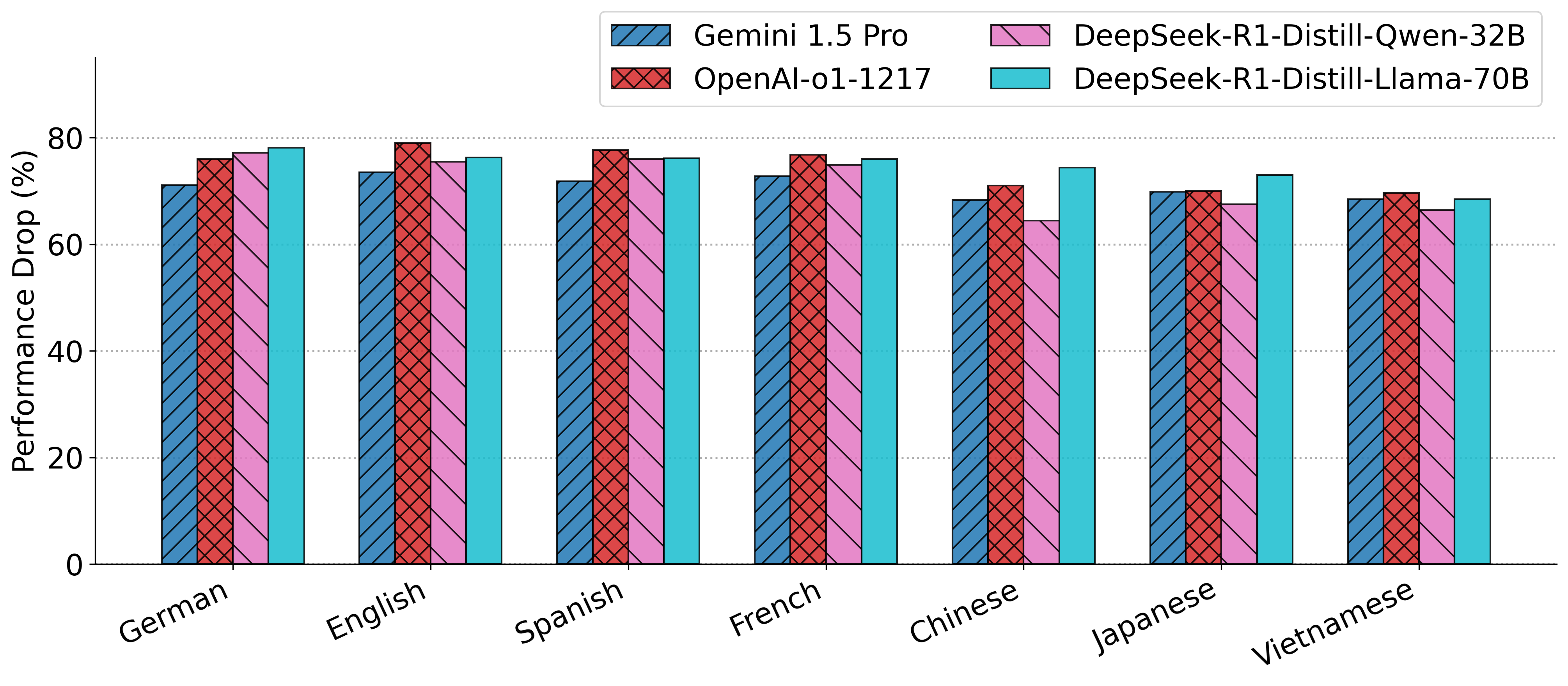}
  \caption{The chart quantifies the performance drop for four \textit{COLA + <model>} on the more challenging MultiSpider 2.0 benchmark, highlighting the impact of linguistic variation.}
  \label{fig:linguistic_variation}
\end{figure}

\sstitle{Performance Drop Across Languages}
\autoref{fig:linguistic_variation} illustrates a substantial performance drop (65--80\%) for all four \textit{COLA-enhanced models} when transitioning to the more challenging \textit{MultiSpider 2.0} benchmark. This sharp decline underscores the increased query complexity and semantic ambiguity present in the newer dataset. While this drop appears marginally smaller for East Asian languages like Chinese, Japanese, and Vietnamese, this is a misleading artifact, as their initial baseline accuracies were already notably lower than those for European languages. This indicates that the performance degradation is a challenge, highlighting the need for improving cross-lingual ability in advanced text-to-SQL models.

\begin{figure}[!htbp]
  \centering
  \includegraphics[width=0.7\linewidth]{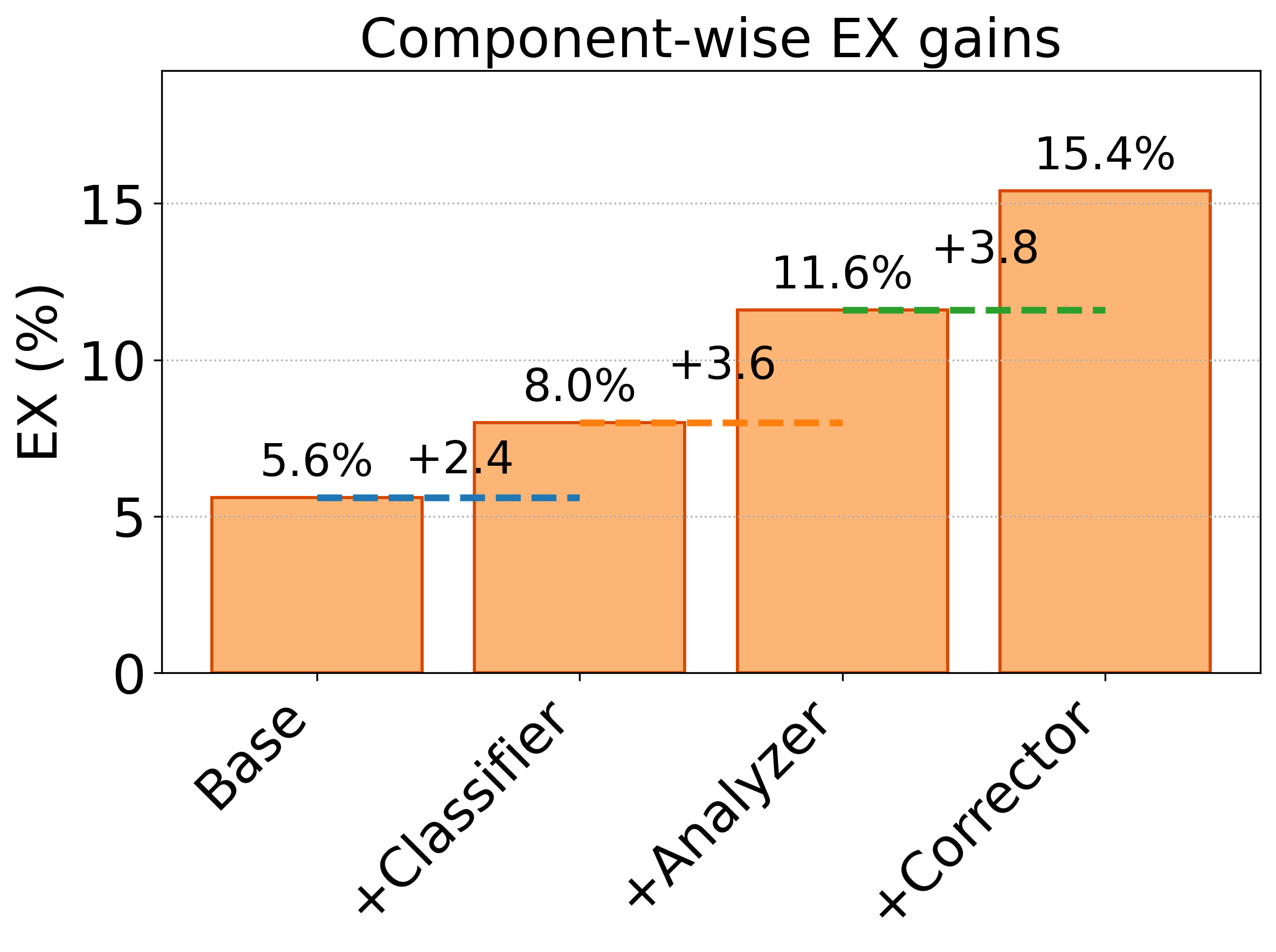}
  \caption{The waterfall chart illustrates the cumulative gains in Execution Accuracy from each module.}
  \vspace{-1em}
  \label{fig:waterfall_ablation}
\end{figure}

\sstitle{Component-wise Contribution to Accuracy}
The ablation study in \autoref{fig:waterfall_ablation} systematically evaluates the contribution of each COLA module, validating our modular design. The analysis shows a clear cumulative effect on execution accuracy. Starting from a \textit{5.6\%} baseline with the core LLM, performance is sequentially boosted by the integration of the \textit{Classifier} (+2.4\%), the \textit{Analyzer} (+3.6\%), and finally the \textit{Corrector} (+3.8\%). This step-wise enhancement brings the final accuracy to \textit{15.4\%}, with \textit{Corrector} emerging as the most impactful component in the framework.

\section{Related Work}

\sstitle{Multilingual Text-to-SQL Datasets}
Progress in Text-to-SQL has been driven by increasingly realistic datasets, moving from single-table supervision (WikiSQL) \cite{zhong2017seq2sql} to cross-domain, multi-table evaluation (Spider) \cite{yu2018spider}. Multilingual efforts—including MultiSpider~1.0 \cite{dou2023multispider}—extend beyond English but often remain limited in scale (one to seven languages), schema coverage, or the extent of localization \cite{min2019pilot,nguyen2020pilot,guo2021chase,jose2021mrat}. In parallel, multilingual task-oriented dialogue corpora scale to six–nine languages and emphasize end-to-end intent/slot modeling rather than SQL grounding \cite{li2020mtop,xu2020end}. This leaves a gap between \textit{multilingual} NLU resources and the schema-grounded reasoning required for database querying. MultiSpider~2.0 addresses this gap with eight languages, richer cross-database coverage, and increased difficulty relative to MultiSpider~1.0 \cite{dou2023multispider}. Beyond a larger language set, the benchmark explicitly targets (i) dialectal and orthographic variation, (ii) robust schema linking in enterprise-style databases, and (iii) compositional SQL (nesting, grouping, windowing), exposing failure modes that small-scale multilingual datasets underplay. By releasing both questions and database content in each language, the evaluation becomes sensitive to lexical and terminological drift rather than only translation artifacts.

\sstitle{Advances in Text-to-SQL Modeling}
Modeling advances largely fall into two families: specialized neural parsers and LLM-based sequence models. Relation-aware encoders that represent schema topology and value mentions improve grounding and column/table selection \cite{wang2019rat}; intermediate SQL representations reduce search complexity and inject structural priors \cite{guo2019towards}; and grammar/type-constrained decoding improves executable correctness by ruling out ill-typed partial programs \cite{yin2018tranx}. Execution-time constraints such as PICARD further stabilize inference with incremental validity checks \cite{scholak2021picard}. In parallel, pretrained seq2seq models (mBART, T5) provide strong multilingual and cross-task priors \cite{liu2020mbart,raffel2020exploring}, narrowing the gap between task-specific parsers and general-purpose LMs. Despite these advances, multilingual Text-to-SQL remains challenging: lexical ambiguity, entity normalization, and schema linking degrade transfer, and English-only gains do not consistently carry over since databases and questions are localized. Our experiments show that increases in language coverage and schema complexity stress both specialized parsers and LLM-based decoders.

\sstitle{Code Generation and Interactive Reasoning}
Text-to-SQL intersects broader code generation and data-science automation—program synthesis from natural language, execution feedback, and tool use \cite{yu2018spider,lin2018nl2bash,chen2021evaluating,lai2023ds,yin2022natural,huang2024code,chan2024mle,Le_Pham_Quan_Luu_2024,pham-etal-2024-unibridge}. Agent-style methods couple reasoning traces with environment interaction to improve faithfulness and problem solving \cite{yao2022react,zhang2023act,chen2023teaching,wang2023plan}. Within Text-to-SQL, both supervised finetuning tailored to schema linking and SQL structure \cite{li2024codes} and carefully engineered prompting pipelines \cite{dong2023c3,wang2023mac,zhang2023act,talaei2024chess,pourreza2024din,gao2023text} are competitive; the latter combine constrained decoding, self-consistency, and retrieval of schema context. Multilingual robustness, however, still hinges on high-quality localization and normalization of database content. Our benchmark complements this literature by providing a multilingual setting in which modeling choices (finetuning vs.\ prompting; single- vs.\ multi-agent) can be stress-tested along axes of language, schema, and compositionality.

\section{Conclusion and Future Work}
MultiSpider~2.0 surfaces challenges that prior benchmarks underplay: enterprise-scale schemas, multilingual and dialectal variation, and deeply compositional SQL. Large, heterogeneous databases with dozens of tables and many-to-many bridges stress long-horizon reasoning and precise schema linking. By coupling high-complexity schemas with multilingual prompts and fine-grained diagnostics, the benchmark exposes performance cliffs that are masked on smaller or monolingual settings and makes clear that language understanding, schema reasoning, and execution must be addressed as a coupled problem rather than isolated modules.

COLA improves reliability over single-LLM reasoning through structured planning, execution-time checks, and selective re-ranking. The framework is modular and budget-aware: plans can be coarser or finer, verification can be staged so that cheap checks run first, and reranking can be truncated to meet latency constraints. Across analyses we repeatedly observe failure modes that recur across models-join-path overreach, spurious aggregation or grouping, entity and alias normalization gaps, and temporal or unit mismatches -- and we show that several of these can be mitigated with targeted planners, schema-graph heuristics, and verifier hooks. Nevertheless, non-English scenarios, especially Chinese, Japanese, and Vietnamese, remain difficult due to linguistic ambiguity, code-switching, and brittle schema linking (e.g., alias/column disambiguation and join-path selection).

Looking ahead, we see several priorities to make multilingual Text-to-SQL robust and deployable:
\begin{itemize}
  \item \textit{Schema-grounded planning}: stronger, stepwise plans that bind entities and relations early, select plausible join paths from the foreign-key graph, and propagate constraints through joins, aggregation, grouping, and nesting.
  \item \textit{Dialect-aware normalization}: canonicalization of entities, units, and date/number formats; support for code-switching, transliteration, and regional synonyms (e.g., en--zh variants), with lightweight lexicons or learned normalizers integrated into the parsing loop.
  \item \textit{Execution-grounded learning}: incorporate process supervision and post-execution feedback, for example verifier-guided search and RLHF~\cite{christiano2017deep,ouyang2022training}, while budgeting test-time compute with early stopping and selective re-ranking~\cite{yang-etal-2022-generating}.
  \item \textit{Multilingual data and adaptation}: targeted fine-tuning with paraphrases, dialectal variants, and counterfactuals; curricula over schema complexity and compositional depth; contrastive pairs that tease apart near-synonyms and subtle quantifier or temporal differences.
\end{itemize}

MultiSpider~2.0 sets a higher, more realistic bar for evaluation and reveals where current methods break. COLA narrows but does not eliminate the gap. We hypothesize that progress will come from tighter coupling of language, schema, and execution, combined with multilingual adaptation and diagnostics that emphasize reliability under distributional shift. Future work may broaden language coverage, including lower-resource varieties, and introduce stress tests for temporal and unit reasoning, code-switching, and schema evolution. Together, these tools support reproducible head-to-head comparisons and move the field from promptable demos toward deployment-grade Text-to-SQL in multilingual environments.

%\section*{Ethical Statement}
%MultiSpider~2.0 is a free, open dataset derived from Spider~2.0 \cite{lei2024spider}. Twenty annotators (translators and NLP researchers) created labels. All questions target open-access databases. Dataset details appear in Section~\ref{sec:multispider2.0}.

% \begin{acks}
% This work was supported by Australian Research Council Discovery Project (Grant No. DP240101108).
% \end{acks}

\balance

% \bibliographystyle{ACM-Reference-Format}
% \bibliography{../ref}

%%% -*-BibTeX-*-
%%% Do NOT edit. File created by BibTeX with style
%%% ACM-Reference-Format-Journals [18-Jan-2012].

\end{document}